\documentclass[journal]{IEEEtran}%

\usepackage{hyperref}       
\usepackage{url}            
\usepackage{booktabs}       
\usepackage{amsfonts}       
\usepackage{nicefrac}       
\usepackage{microtype}      
\usepackage{lipsum}
\usepackage{caption}
\usepackage{graphicx}
\usepackage{subfigure}
\usepackage{color}
\usepackage{multirow}
\usepackage{amsmath}
\usepackage{cite}
\usepackage{xcolor}
\usepackage{bm}
\usepackage{array}
\usepackage{tablefootnote}

\begin{document}
\title{Rectified Linear Postsynaptic Potential Function for Backpropagation in Deep Spiking Neural Networks}
\author{Malu~Zhang,
        Jiadong~Wang,
        Burin~Amornpaisannon,
        Zhixuan~Zhang,
        VPK Miriyala,
        Ammar~Belatreche,
        Hong~Qu,
        Jibin~Wu,
        Yansong Chua,
        Trevor E. Carlson,~\IEEEmembership{Senior~Member,~IEEE,}
        and~Haizhou~Li,~\IEEEmembership{Fellow,~IEEE}
\thanks{This work has been submitted to the IEEE for possible publication.
Copyright may be transferred without notice, after which this version may
no longer be accessible.}
\thanks{
This research is supported by Programmatic Grant No. A1687b0033 from the Singapore Government's Research, Innovation and Enterprise 2020 plan (Advanced Manufacturing and Engineering domain), the National Key Research and Development Program of China (2018AAA0100202), the Zhejiang Lab (Grant No. 2019KC0AB02).}

\thanks{M. Zhang, J. Wang, J. Wu, Y. Chua and H. Li are with the Department of Electrical and Computer Engineering, National University of Singapore, Singapore.}%
\thanks{B. Amornpaisannon, V. P. K. Miriyala, and T. Carlson are with the School of Computing, Department of Computer Science, National University of Singapore, Singapore.}%
\thanks{Z. Zhang, and H. Qu are with the School of Computer
	Science and Engineering, University of Electronic Science and Technology
	of China, Chengdu 610054, China.}
\thanks{A. Belatreche is with the Department of Computer and Information
	Sciences, Faculty of Engineering and Environment, Northumbria
	University, Newcastle upon Tyne NE1 8ST, U.K.}
\iffalse \thanks{\emph{M. Zhang, J. Wang, B. Amornpaisannon and Z. Zhang contributed equally to this work.}}\fi}
\maketitle
\begin{abstract}
Spiking Neural Networks (SNNs) use spatio-temporal spike patterns to represent and transmit information, which is not only biologically realistic but also suitable for ultra-low-power event-driven neuromorphic implementation. Motivated by the success of deep learning, the study of Deep Spiking Neural Networks (DeepSNNs) provides  promising directions for artificial intelligence applications. However,  training of DeepSNNs is not straightforward because the well-studied error back-propagation (BP) algorithm is not directly applicable. In this paper, we first establish an understanding as to why error back-propagation does not work well in DeepSNNs. To address this problem, we propose a simple yet efficient Rectified Linear Postsynaptic Potential function (ReL-PSP) for spiking neurons and propose a Spike-Timing-Dependent Back-Propagation (STDBP) learning algorithm for DeepSNNs. In STDBP algorithm, the timing of individual spikes is used to convey information (temporal coding), and learning (back-propagation) is performed based on spike timing in an event-driven manner. Our experimental results show that the proposed learning algorithm achieves state-of-the-art classification accuracy in single spike time based learning algorithms of DeepSNNs. Furthermore, by utilizing the trained model parameters obtained from the proposed STDBP learning algorithm, we demonstrate the ultra-low-power inference operations on a recently proposed neuromorphic inference accelerator. Experimental results show that the neuromorphic hardware consumes 0.751~mW of the total power consumption and achieves a low latency of 47.71~ms to classify an image from the MNIST dataset. Overall, this work investigates the contribution of spike timing dynamics  to information encoding, synaptic plasticity and decision making, providing a new perspective to design of future DeepSNNs and neuromorphic hardware systems.
\end{abstract}


\section{Introduction}
\IEEEPARstart{T}{he} success of Deep Neural Networks (DNNs) is attributed to their underlying deep hierarchical structure that learns the  representations of big data with multiple levels of abstraction~\cite{lecun2015deep}. With DNNs, we have advanced the state-of-the-art in many machine learning tasks by leaps and bounds, such as image recognition\cite{he2016deep}, speech recognition\cite{abdel2014convolutional,lam2019gaussian}, natural language processing \cite{young2018recent,ghaeini2018dr,lee-li-2020-modeling,lee2019linguistically}, and medical diagnosis \cite{esteva2017dermatologist}. However, training of DNNs generally requires high computing resources (e.g., GPUs and computing clusters). Therefore, in power-critical computing platforms, such as edge computing, the implementation of DNNs is drastically limited \cite{feldmann2019all,shrestha2018slayer}. Spiking Neural Networks (SNNs) provide a low-power alternative to neural network implementation. It is designed to emulate brain computing, that has the potential to provide computing capabilities equivalent to that of DNNs on an ultra-low-power spike-driven neuromorphic hardware \cite{izhikevich2003simple,gerstner2002spiking,pfeiffer2018deep,li2017liquid,zhang2019fast}. However, due to the relatively shallow network structures, SNNs have yet to match the performance of their DNN counterparts in pattern classification tasks on standard benchmarks\cite{deng2020rethinking}. 

There has been a growing interest in the implementation of deep structures for SNNs (DeepSNNs)
\cite{shrestha2018slayer,pfeiffer2018deep,pei2019towards,tavanaei2018deep,wu2018spatio,wu2019direct,jin2018hybrid,wu2019hybrid,wu2019deep}. Unfortunately, the training of DeepSNNs is not straightforward as the well-studied error back-propagation (BP) learning algorithm is not directly applicable due to the complex temporal dynamics and the non-differentiable spike function. Addressing the issue of DeepSNN training,  successful implementations can be grouped into three categories.

The first category is ANN-to-SNN conversion methods, where we first train an ANN, then approximate the pre-trained ANN with an SNN equivalent~\cite{esser2015backpropagation,hunsberger2015spiking,essera2016convolutional,o2013real,liu2017noisy,diehl2015fast,diehl2016truehappiness,rueckauer2017conversion,rueckauer2018conversion,han2020rmp}. The idea of such ANN-to-SNN approximation  benefits from the state-of-the-art ANN training algorithms and uses an equivalent SNN  to approximate the performance of the pre-trained ANN. However, the conversion suffers from loss of accuracy due to the approximation. Despite much progress, such as weight/activation normalization \cite{diehl2015fast,diehl2016truehappiness,rueckauer2017conversion}, and adding noise to the model \cite{o2013real,liu2017noisy}, the solutions are far from being perfect. Furthermore, as the spiking rate of a neuron is typically used to encode analog activity of a classical artificial neuron, a high level of activation requires a high spike rate, which then hides the timing information of the spiking activity, and is energy-intensive \cite{mostafa2017supervised}. Furthermore,  the inference time of the spike rate based encoding scheme is another problem~\cite{wu2019hybrid}.

The second category is the membrane potential driven learning algorithms, treating the neuron's membrane potential as differentiable signals which are used to solve the non-differentiable problems of spikes with using surrogate derivatives \cite{neftci2019surrogate}. For example, \cite{panda2016unsupervised,lee2016training,zenke2018superspike} back-propagate errors based on the membrane potential at a single time step, which ignores the temporal dependency, but only use signal values at the current time instance. To address this problem, SLAYER \cite{shrestha2018slayer} and STBP \cite{wu2018spatio,wu2019direct} train DeepSNNs with surrogate derivatives based on the idea of Back-propagation Through Time (BPTT) algorithm. While competitive accuracies are reported on the MNIST and CIFAR10 datasets \cite{wu2019direct}, the computational and memory requirements of these algorithms are high for BPTT because the entire sequence must be stored to compute the gradients accurately.

The third category is the spike-driven learning algorithms, which uses the timing of spikes as the relevant signals for controlling synaptic changes. The typical examples include SpikeProp \cite{bohte2002error} and its derivatives \cite{shrestha2017robust,xu2013supervised,robustnessshrestha2017robustness,hong2019training}. These methods apply a linear assumption that the neuron's membrane potential increases linearly in the infinitesimal time around the spike time, allowing the calculation of the derivative of the spike function and facilitating the implementation of back-propagation in multi-layer SNNs. Recently, Mostafa \cite{mostafa2017supervised} applied non-leaky integrate-and-fire neurons to avoid the problem of the non-differentiable spike function, and the work showed competitive performance on MNIST dataset. The performance of SNN with spike-driven learning algorithm is further improved by \cite{kheradpisheh2019s4nn} and \cite{comsa2019temporal}. However, the existing spike-driven learning algorithms suffer from certain limitations, such as the problems of dead neurons and gradient exploding \cite{mostafa2017supervised}. For example, in \cite{mostafa2017supervised}, some constraints are imposed on the synaptic weights to overcome the dead neuron problem, and a gradient normalization strategy is used to overcome the problem of gradient exploding. These complex training strategies limit the scalability of these learning algorithms. 

Among the existing learning algorithms, the spike-driven learning algorithms perform the SNNs training in a strictly event-driven manner, and are compatible with the temporal coding in which the information is carried by the timing of individual spikes in a very sparse manner; Hence, spike-driven learning algorithms hold the potential of enabling training and inference in low-power devices. Recently, several neuromorphic architectures have been proposed for accelerating the SNN inference and training efficiently~\cite{loihi_2018, akopyan_truenorth:_2015, minitaur_2014, esser2015backpropagation, wang2019shenjing, khan2008spinnaker, ji2018bridge}. Currently, the most prominent among them is Intel's Loihi~\cite{loihi_2018}, which can support on-chip learning with a wide range of spike-timing-dependent-plasticity (STDP) rules. During SNN acceleration, Loihi is claimed to be 1000$\times$ faster than the general-purpose processors such as CPUs and GPUs, while using much less power. In addition, Srivatsa et al, \cite{srivatsa2020} recently proposed a neuromorphic inference accelerator called YOSO. YOSO can significantly leverage the sparse spiking activity in SNNs and facilitates the ultra-low-power inference operations with high classification accuracies \cite{srivatsa2020}. With these considerations, we develop an effective spike-driven learning algorithms for DeepSNNs with the  following contributions in this paper:

1) We thoroughly analyze the issues that make the well-studied BP algorithm incompatible for training SNNs, including the problems of non-differentiable spike generation function, gradient exploding and dead neuron. Building on such an understanding, we put forward a Rectified Linear Postsynaptic Potential function (ReL-PSP) for spiking neurons to resolve these problems. 

2) Based on the proposed ReL-PSP, we derive a new spike-timing-dependent BP algorithm (STDBP) for DeepSNNs. In this algorithm, the timing of spikes is used as the information-carrying quantities, and learning happens only at the spike times in a totally event-driven manner.

3) As STDBP with ReL-PSP is highly scalable, we extend it to convolutional spiking neural network (C-SNN), and achieve an accuracy of 99.4\% on the MNIST dataset. To the best of our knowledge, this is the first implementation of a C-SNN structure in the area of single-spike-timing-based supervised learning algorithms, with a state-of-the-art accuracy performance. 

4) Lastly, using the weights obtained through the proposed STDBP learning algorithm, we demonstrate the ultra-low-power inference operations by accelerating our SNN model on the neuromorphic inference accelerator, YOSO. 

Overall, this work not only provides a novel perspective to investigate the significance of spike timing dynamics in information coding, synaptic plasticity, and decision making in SNNs-based computing paradigm, but also demonstrates the possibility of realizing low power inference operations with high classification accuracy.

\section{Problem Analysis}
\label{problem description}
\begin{figure*}[h]
\centering
\includegraphics[scale=0.55]{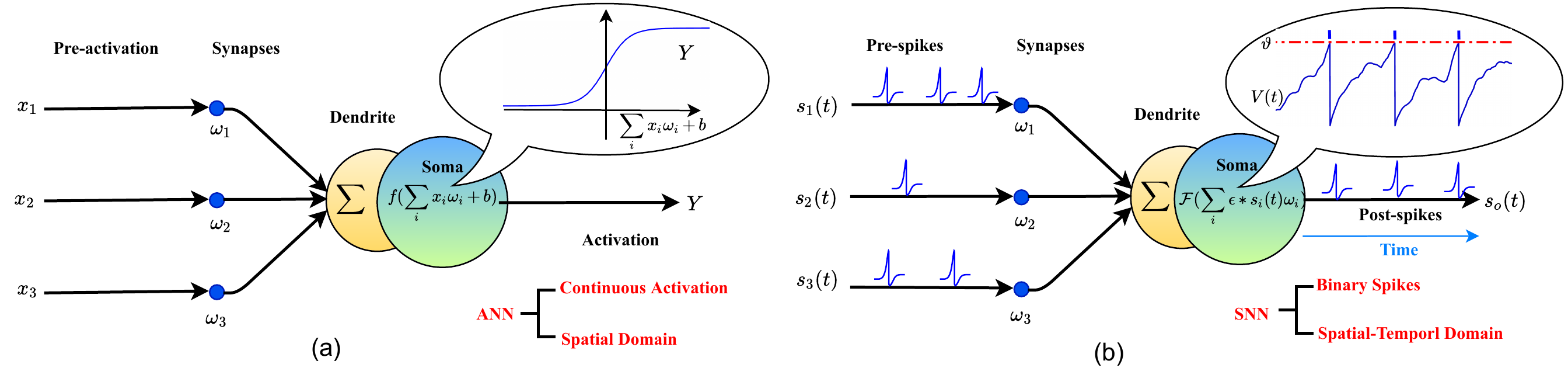}
    \caption{Neuron models in DNNs and SNNs. (a) A typical DNN neuron model, in which the information from previous layer arrives in the form of real values in the spatial domain. $x$, $\omega$, $b$, and $Y$ are input activation, synaptic weights, bias and output activation, respectively. Output $Y$ is produced by the differentiable activation function $f(\cdot)$. (b) A typical spiking neuron model, in which information from previous layer arrives in the form of spatial-temporally distributed spike events. $s_i(t)$, $\omega$, and $s_o(t)$ are input spikes, synaptic weights and output spikes, respectively. The non-differentiable function $\mathcal{F}(\cdot)$ generates output spikes $s_o(t)$ from the membrane potential $V(t)$.}
\label{Fig0}
\end{figure*}

Error back-propagation (specifically stochastic gradient descent) is the workhorse for the remarkable success of DNNs. However, as shown in Fig.~\ref{Fig0}, the dynamics of a typical artificial neuron in DNNs and that in SNNs is rather different, therefore the well-studied BP algorithm cannot be directly applied to DeepSNNs due to non-differentiable spike function, exploding gradients, and dead neurons. In the following, we will discuss these issues in depth.

Consider a fully connected DeepSNN. For simplicity, each neuron is assumed to emit at most one spike. In general, the membrane potential $V_j^l$ of neuron $j$ in layer $l$ with $N$ presynaptic connections can be expressed as,
 
 \begin{equation}
V^l_j(t)=\sum_i^N\omega_{ij}^{l}\varepsilon(t-t_i^{l-1})-\eta(t-t^l_{j})
\label{old neuronmodel}
\end{equation}
where $t_i^{l-1}$ is the spike of the $i$th neuron in layer $l$-$1$, and $\omega^{l}_{ij}$ is the synaptic weight of the connection from neuron $i$ (in $l$-1 layer) to neuron $j$ (in $l$ layer). 
Each incoming spike from neuron $i$ will induce a postsynaptic potential (PSP) at neuron $j$, and the kernel $\varepsilon(t-t_i^{l-1})$ is used to describe the PSP generated by the spike $t_i^{l-1}$. Hence each input spike makes a contribution to the membrane potential of the neuron as described by $\omega_{ij}^{l}\varepsilon(t-t_i^{l-1})$ in Eq. \ref{old neuronmodel}. There are several PSP functions, and a commonly used one is  alpha function which is defined as
\begin{equation}
    \varepsilon(t)=\frac{t}{\tau} \text{exp} \big(1-\frac{t}{\tau}\big)   \quad\quad\quad\quad\quad t>0
    \label{psp}
\end{equation}

\begin{figure*}[h]
\centering
\includegraphics[scale=0.9]{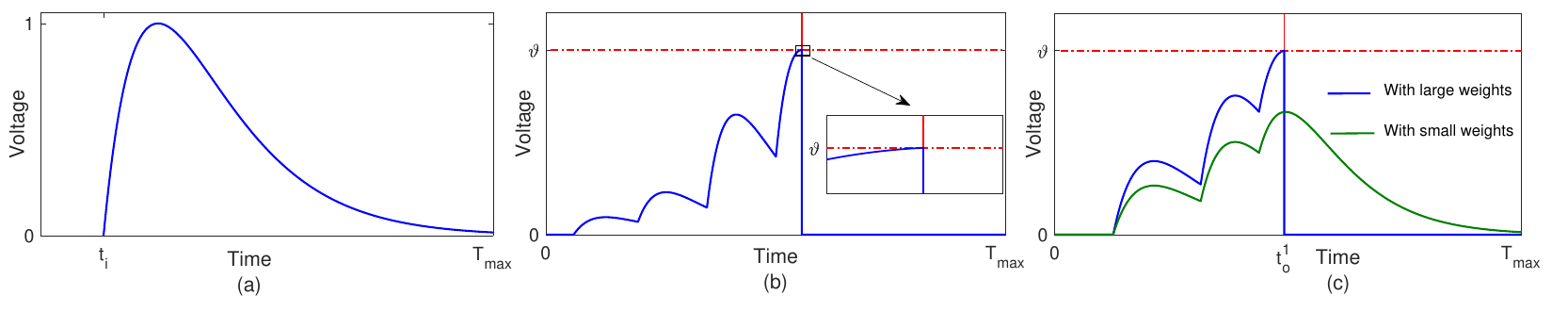}
\caption{(a) Alpha shape PSP function. (b) The membrane potential barely reaches the firing threshold, and exploding gradient occurs. (c) The alpha-PSP neuron with weak synaptic weights is susceptible to be a dead neuron .}
\label{Fig1}
\end{figure*}

Fig. \ref{Fig1}(a) shows the alpha-PSP response function. As shown in Fig. \ref{Fig1}(b), integrating the weighted PSPs gives the dynamics of the membrane potential $V_j^l(t)$. The neuron $j$ will emit a spike when its membrane potential $V_j^l(t)$ reaches the firing threshold $\vartheta$, as mathematically defined in the spike generation function $\mathcal{F}$:
\begin{equation}
    t_j^l = \mathcal{F} \left\{ t |V_j^l(t)=\vartheta, t \geq 0 \right\}
    \label{spikegenerate}
\end{equation}
 Once a spike is emitted, the refractory kernel $\eta(t-t^l_{j})$ is used to reset the membrane potential to resting. 

To train SNNs using BP, we need to compute the derivative of the postsynaptic spike time $t^l_j$ with respect to a presynaptic spike time $t^{l-1}_i$ and synaptic weight $\omega^l_{ij}$ of the corresponding connection:

\begin{equation}
   \frac{\partial t_j^l}{\partial \omega_{ij}^l}=\frac{\partial t_j^l}{\partial V_j^l{(t_j^l)}}\frac{\partial V_j^l(t_j^l)}{\partial \omega_{ij}^{l}} \quad\quad\text{if}\quad\quad  t_j^l>t_i^{l-1}
   \label{tw1}
\end{equation}

\begin{equation}
   \frac{\partial t_j^l}{\partial t_i^{l-1}}=\frac{\partial t_j^l}{\partial V_j^l{(t_j^l)}} \frac{\partial V_j^l{(t_j^l)}}{\partial t_i^{l-1}} \quad\quad\text{if}\quad\quad  t_j^l>t_i^{l-1}
   \label{tt1}
\end{equation}

 Due to the discrete nature of the spike generation function (Eq. \ref{spikegenerate}), the difficulty of Eq. \ref{tw1} lies in solving the partial derivative ${\partial t_j^l}/{\partial V_j^l{(t_j^l)}}$ , which we refer to as the problem of \textbf{non-differentiable spike function}. Existing spike-driven learning algorithms \cite{bohte2002error,yu2018spike} assume that the membrane potential $V^l_j(t)$ increases linearly in the infinitesimal time interval before spike time $t_j$. Then, ${\partial t_j}/{V_j(t)}$ can be expressed as 
\begin{equation}
\frac{\partial t_j^l}{\partial V_j^l{(t_j^l)}}=\frac{-1}{\partial V_j^l{(t_j^l)}/\partial t_j^l}=\frac{-1}{\sum_i^N\omega^l_{ij}\frac{\partial \varepsilon(t_j^l-t_i^{l-1})}{\partial t_j^l}}
    \label{2err1}
\end{equation}
with 
\begin{equation}
\frac{\partial \varepsilon(t_j^l-t_i^{l-1})}{\partial t_j^l}=\frac{\text{exp}(1-(t_j^l-t_i^{l-1})/\tau)}{\tau^2}(\tau+t_i^{l-1}-t_j^l)
\label{2err2}
\end{equation}

The \textbf{exploding gradient} problem occurs when $\partial V_j^l{(t_j^l)}/\partial t_j^l\approx0$ i.e. the membrane potential is reaching the firing threshold, emitting a spike (Fig. \ref{Fig1}b). Since $\partial V_j(t_j)/\partial t_j$ is the denominator in Eq. \ref{2err1}, this causes Eq. \ref{2err1} to explode with large weight updates. Despite the progress, such as adaptive learning rate \cite{shrestha2015adaptive} and dynamic firing threshold \cite{hong2019training}, the problem has not been fully resolved. 

From Eq. \ref{tw1} and Eq. \ref{tt1}, when the presynaptic neuron does not emit a spike, the error cannot be back propagated through $\partial V_j^l(t^l_j)/\partial t_i^{l-1}$. This is the \textbf{dead neuron} problem. This problem also exists with analog neurons with ReLU activation function in DNNs. However, due to the leaky nature of the PSP kernel and spike generate mechanism, spiking neurons encounter a more serious dead neuron problem. As shown in Fig.~\ref{Fig1}(c), there are three input spikes, and the neuron emits a spike with large synaptic weights (blue). With slightly reduced synaptic weights, the membrane potential stays sub-threshold and the neuron becomes a dead neuron (green). When the neuron does not spike, no errors can back-propagate through it. The problem of dead neuron is fatal in spike-driven learning algorithms.

\section{Spiking Neuron Model and Learning Algorithm}
\label{sec3}
In this section, we describe how the above challenges may be overcome, thereby a DeepSNN may still be trained using BP. To this end, we introduce Rectified Linear Postsynaptic Potential function (ReL-PSP) as a new spiking neuron model, and present a spike-timing-dependent back propagation (STDBP) learning algorithm that is based on ReL-PSP.

\subsection{ReL-PSP based spiking neuron model}

\label{sec3A}
\begin{figure*}[h]
\centering
\includegraphics[scale=0.9]{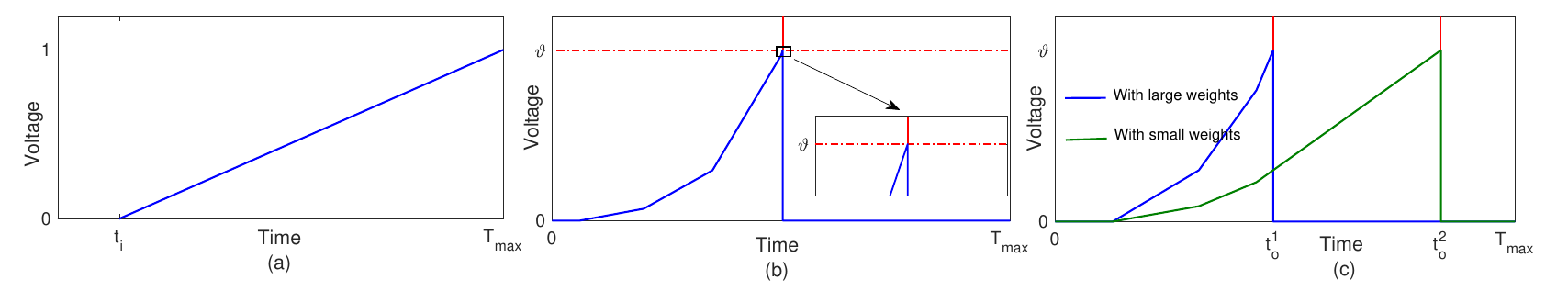}
\caption{There are three input spikes denoted as $t_1,t_2,t_3$. The blue and green lines show the membrane potential with large and small synaptic weights, respectively. (a) ReL-PSP function. (b) Trace of the neuron membrane potential during threshold crossing. (c) The ReL-PSP neuron generates spikes at $t_o^1$ and $t_o^2$ with large and small synaptic weights, respectively ($t_o^1<t_o^2$).}
\label{Fig2}
\end{figure*}

As presented in Section \ref{problem description}, BP cannot be directly applied in DeepSNNs due to problems of non-differentable spike function, exploding gradient and dead neuron. To overcome the problems, we propose a simple yet efficient Rectified Linear Postsynaptic Potential (ReL-PSP) based spiking neuron model,  whose dynamics is defined as follows,
\begin{equation}
V^l_j(t)=\sum_i^N\omega_{ij}^{l}K(t-t_i^{l-1})
\label{neuronmodel}
\end{equation}
whereby $K(t-t_i^{l-1})$ is the kernel of the PSP function, which is defined as 
\begin{equation}K(t-t_i^{l-1}) =
\begin{cases}
t-t_i^{l-1} \quad \text{if} \quad t>t_i^{l-1}\\ 
0 \quad\quad\quad\quad\text{otherwise}
\end{cases}
\label{ReL-PSP}
\end{equation}
As shown in Fig. \ref{Fig2}(a), given an input spike at $t_i^{l-1}$, the membrane potential after $t_i^{l-1}$ is a linear function of time $t$. Since the shape of the proposed PSP function resembles that of a rectified linear function, we name it the ReL-PSP function. In the following, we will analyze how the proposed neuron model solves the above-mentioned problems.

\subsubsection{Non-differentiable spike function}
As shown in Fig. \ref{Fig2}b, due to the linearity of the ReL-PSP, the membrane potential $V_j^l(t)$ increases linearly prior to spike time $t_j^l$. The linearity is a much desired property from a postsynaptic potential function. We can now directly use Eq. \ref{tjvj} to compute ${\partial t_j^l}/{\partial V_j^l{(t_j^l)}}$. This resolves the issue of non-differentiable spike generation. 

\begin{equation}
\begin{split}
\frac{\partial t_j^l}{\partial V_j^l{(t_j^l)}}&=-\frac{1}{{\partial V_j^l{(t_j^l)}}/{\partial t_j^l}}\\
&= \frac{-1}{\sum_i^N \omega^l_{ij}\frac{\partial K(t_j^l-t_i^{l-1})}{\partial t_j^l}}\\
&=\frac{-1}{\sum_i^N \omega^l_{ij}}\quad\quad\text{if}\quad  t_j^l>t_i^{l-1}
   \label{tjvj}
\end{split}
\end{equation}

The precise gradients in BP provide the necessary information for network optimization, which is the key to the performance of DNNs. Without having to assume linearity, we use the precise value of ${\partial t_j^l}/{\partial V_j^l{(t_j^l)}}$ instead of approximating it, and avoid accumulating errors across multiple layers.

\subsubsection{Gradient explosion}
Exploding gradient occurs when the denominator in Eq. \ref{2err1} approaches 0. In this case, the membrane potential just reaches the firing threshold at spike time, and is caused by the combined effect of $\omega_{ij}^l$ and partial derivative of the PSP function. Alpha-PSP function has zero gradient at its peak. It is clear that  ReL-PSP function has a lower chance of zero gradient than alpha-PSP function. As the partial derivative of ReL-PSP, $\partial K(t_j^l-t_i^{l-1})/\partial t_j^l$, is always equal to 1, Eq. \ref{tjvj} can be expressed as ${-1}/{\sum_i^N \omega^l_{ij}}$.

As $\sum_i^N \omega^l_{ij}$ may still be close to $0$, the exploding gradient problem may not be completely solved. However, from Eqs. \ref{spikegenerate} and \ref{neuronmodel}, we obtain the spike time $t_j^l$ as a function of input spikes and synaptic weights $ \omega^l_{ij}$, and the spike time $t_j^l$ can be calculated as,
\begin{equation}
t_j^l=\frac{\vartheta+\sum_i^N\omega_{ij}^{l}t_i^{l-1}}{\sum_i^N\omega_{ij}^{l}}
\label{wt}
\end{equation}

Should the $\sum_i^N\omega_{ij}^{l}$ be close to 0, the spike $t_j^l$ will be emitted late, and may not contribute to the spike  $t_j^{l+1}$ in the next layer. Therefore, the neuron $j$ in the $l$ layer does not participate in error BP, and does not result in exploding gradient. 

\subsubsection{Dead neuron}
In neural networks, sparse representation (few activated neurons) has many advantages, such as information disentangling, efficient variable-size representation, linear separability etc. However, sparsity may also adversely affect the predictive performance. Given the same number of neurons, sparsity reduces effective capacity of the model \cite{glorot2011deep}.
Unfortunately, as shown in Fig. \ref{Fig1}(c), due to the leaky nature of the alpha-PSP and the spike generation mechanism, such a spiking neuron is more likely to suffer from the dead neuron problem. 

As shown in Fig. \ref{Fig2}(c), with the ReL-PSP kernel, the PSP increases over time within the simulation window $T_{max}$. Hence the neuron with a more positive sum of weights fires earlier than one with a less positive sum, with lower probability of becoming a dead neuron. Overall, the proposed ReL-PSP greatly alleviates the dead neuron problem as the PSP does not decay over time, while maintaining a sparse representation to the same extent of the ReLU activation function.

\subsection{Error backpropagation}
\label{sec:error_bp}
Given a classification task with $n$ categories, each neuron in the output layer is assigned to a category. When a training sample is presented to the neural network, the corresponding output neuron should fire the earliest. Several loss functions can be constructed to achieve this goal \cite{mostafa2017supervised,kheradpisheh2019s4nn,comsa2019temporal}. In this work, the cross-entropy loss function is used. To minimise the spike time of the target neuron, at the same time,   maximise the spike time of non-target neurons, we use the softmax function on the negative values of the spike times in the output layer: $p_j=\text{exp}(-t_j)/\sum_i^n \text{exp}(-t_i)$. The loss function is given by,  
\begin{equation}
L(g,\bold{t^o})=-\text{ln}\frac{\text{exp}(-\bold{t^o}[g]))}{\sum_i^n\text{exp}(\bold{-t^o}[i])}
\label{lossfunction}
\end{equation}
where $\bold{t^o}$ is the vector of the spike times in the output layer and $g$ is the target class index\cite{mostafa2017supervised}.

The loss function is minimised by updating the synaptic weights across the network. This has the effect of delaying or advancing spike times across the network. The derivatives of the first spike time $t_j^l$ with respect to synaptic weights $\omega_{ij}^l$ and input spike times $t_i^{l-1}$ are given by 

\begin{equation}
   \frac{\partial t_j^l}{\partial \omega_{ij}^l}=\frac{\partial t_j^l}{\partial V_j^l{(t_j^l)}}\frac{\partial V_j^l(t_j^l)}{\partial \omega_{ij}^{l}}= \frac{ t_i^{l-1}-t_j^l}{\sum_i^N \omega^l_{ij}} \quad\text{if}\quad  t_j^l>t_i^{l-1}
   \label{tw}
\end{equation}

\begin{equation}
   \frac{\partial t_j^l}{\partial t_i^{l-1}}=\frac{\partial t_j^l}{\partial V_j^l{(t_j^l)}} \frac{\partial V_j^l{(t_j^l)}}{\partial t_i^{l-1}}=\frac{ \omega^{l}_{ij} }{\sum_i^N \omega^l_{ij}} \quad\text{if}\quad t_j^l>t_i^{l-1}
   \label{tt}
\end{equation}

Following Eq.~\ref{tw} and Eq.~\ref{tt}, a standard BP can be applied  for DeepSNNs training.

\section{Hardware Model}
\begin{figure*}[t]
\centering
\includegraphics[scale=0.280]{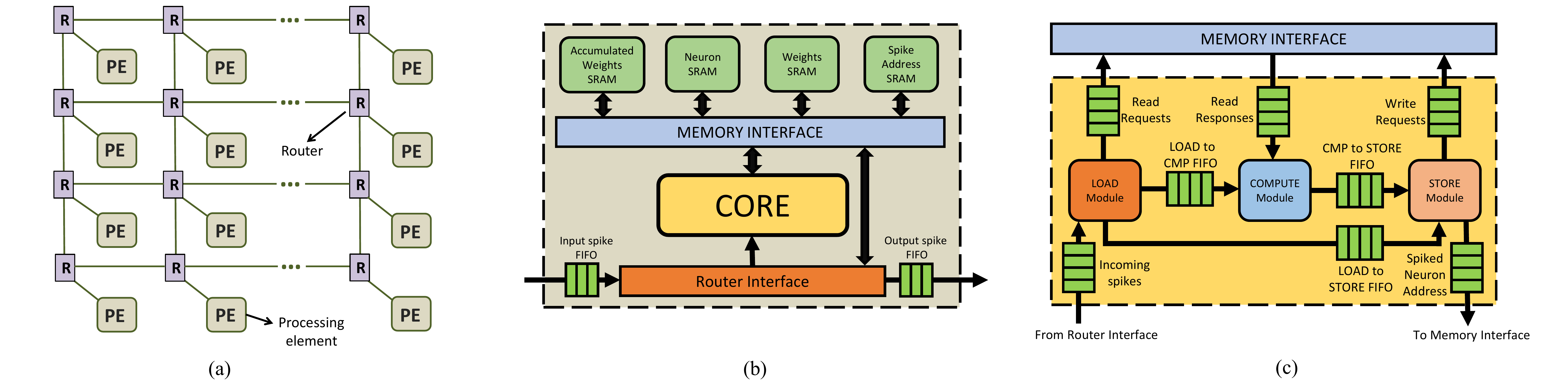}
\caption{(a) The YOSO accelerator with network-on-chip (NoC) architecture, (b) the architecture of each processing element (PE) in YOSO, and (c) the computational core in each PE \cite{srivatsa2020}.}
\label{fig:full_acc}
\end{figure*}

To put the proposed spiking neuron model and STDBP learning algorithm into action, we evaluate them on a hardware architecture for pattern classification tasks. We first implement them in spiking neural networks for inference operations on YOSO platform~\cite{srivatsa2020}. YOSO is specifically designed for accelerating temporal coding based SNN models with sparse spiking activity. It was shown that YOSO facilitates ultra-low-power inference operations ($<$ 1~mW) as a neuromorphic accelerator with state-of-the-art performance.

In this section, we will describe the hardware architecture of YOSO, the technique that maps SNNs on YOSO, and the simulation methodology that evaluates the YOSO performance during inference operations.

\subsection{Hardware Architecture}
\label{harc}

As shown in Fig.~\ref{fig:full_acc} (a), YOSO is a Network-On-Chip (NoC) architecture with multiple processing elements (PEs) connected in a mesh topology. Each PE has a router, which facilitates the propagation of spikes from one PE to another. As shown in Fig.~\ref{fig:full_acc} (b), each PE in the NoC consists of four static random-access memories (SRAMs), a memory interface, a core, a router interface, first-in-first-out (FIFO) buffers to facilitate communication between router and router interface. The router sends or receives the spikes to or from other PEs in the NoC. During the initialization phase, i.e. when the accelerator is downloading the model to be run, the router sends the spikes to SRAMs directly via router and memory interfaces. During the inference phase, the router sends the spikes to the core, where all the computations take place. Note that YOSO~\cite{srivatsa2020} can only accelerate inference operations and training needs to be performed offline. 

The SRAMs are used to store all the information required for processing the incoming spikes and for generating the output spikes. As shown in Fig.~\ref{fig:full_acc} (b), the SRAMs can communicate with the core and router interface via a memory interface. The accumulated weight, neuron, weight, and spike address SRAMs store accumulated weights, neuron potentials, weights between two layers, and the spike addresses of the neurons allocated for that PE, respectively. 

As shown in Fig.~\ref{fig:full_acc} (c), the core consists of three modules--load, compute, and store.  The load module is responsible for decoding the information encoded in incoming spikes. A spike processing algorithm introduced in~\cite{srivatsa2020} is used to encode the information in spike packets that can read by the load module. After decoding the information in incoming spikes, the load module generates read requests to different SRAM blocks. After sending the read requests generated by one incoming spike, it transits to the idle state and waits for the next input spike.  The compute module receives the data requested by the load module from the SRAMs, updates the data using the saturated adder, and sends the results to the FIFO connected to the store module. The store module receives addresses of the loaded data from the load module and the results from the compute module. Using the data and addresses received, it writes the data back to the SRAMs. It also generates spikes when a condition, depending on the chosen techniques, TTFS or softmax, is met to be sent to the next layer. In addition, the core is designed based on the decoupled access-execute model~\cite{smith1982decoupled}, which enables it to hide memory access latency by performing different computations parallelly. 

\subsection{Mapping}
\label{mapping}
In this work, we plan to accelerate our trained SNN models on YOSO. To map a $m \times n$ fully connected SNN layer on YOSO, a minimum of $C = MAX(\frac{n}{N},\frac{m \times n}{W})$ PEs are needed where $N$ is the maximum number of neurons that can be mapped to a single core and $W$ is the maximum number of weights that the core can contain. The PEs are placed within a $\sqrt{C}$ by $\sqrt{C}$ grid. Each PE in the grid will be allocated to a layer or a part of layer for processing received by that layer.

\subsection{Hardware Simulation Methodology}
\label{hsimmeth}
We implemented the YOSO's NoC architecture using the OpenSMART NoC generator~\cite{kwon2017opensmart}. The X-Y routing mechanism is used to send the spike packets from one PE to another. In addition, we designed the PEs in such a way that each PE can meet the memory and computational requirements of at least 256 neurons~\cite{srivatsa2020}. To evaluate the performance of YOSO, we synthesized the hardware architecture shown in Fig.~\ref{fig:full_acc} using Synopsys Design Compiler version P-2019.03-SP5 targeting a 22nm technology node with a $6\times7$ PE configuration. Gate-level simulations are performed using the Synopsys VCS-MX K-2015.09-SP2-9 and power analysis is performed using the Synopsys PrimePower version P-2019.03-SP5. 

\section{Experiments}

In this section, we evaluate two SNNs: a fully connected SNN and a convolutional SNN on image classification task based on the MNIST dataset~\cite{lecun1998gradient}, Fashion-MNIST dataset~\cite{xiao2017fashion} and Caltech 101 face/motorbike dataset\footnote{http:
//www.vision.caltech.edu}. We benchmark their learning capabilities against existing spike-driven learning algorithms. In addition, we evaluate the power and energy efficiency of our fully connected SNN by accelerating its inference operations on YOSO.
\subsection{Temporal coding}

We employ an efficient temporal coding scheme that encodes information as spiking time, with the assumption that strongly activated neurons tend to fire earlier~\cite{thorpe2001spike}. 
 The input information is encoded in spike timing of neurons, with each neuron firing only once. More salient information is encoded as an earlier spike in the corresponding input neuron. The encoding spikes then propagate to  subsequent layers in a temporal fashion. Each neuron in the hidden and output layer receives the spikes from its presynaptic neurons, and emits a spike when the membrane potential reaches a threshold. Similar to the input layer, the neurons in the hidden and output layer that are strongly activated will fire first. Therefore, temporal coding is maintained throughout the DeepSNN, and the output neuron that fires earliest categorizes the input stimulus.

\begin{figure*}
    \centering
    \includegraphics[scale=0.38]{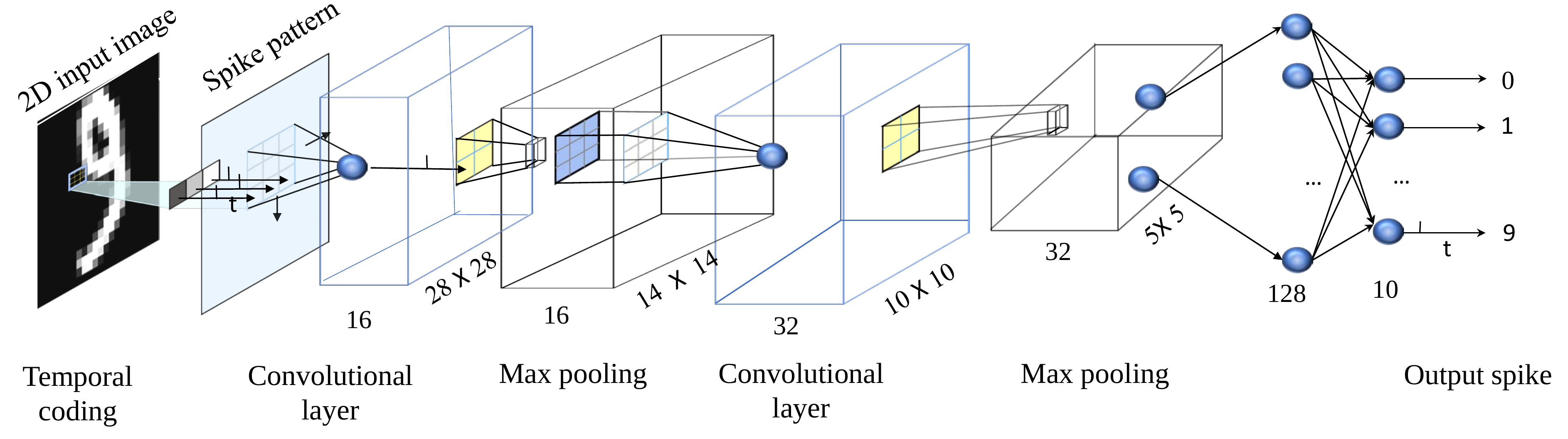}
    \caption{The convolutional spiking neural network.  }
    \label{CNN model}
\end{figure*}
\subsection{MNIST Dataset}

The MNIST dataset comprises of 60,000 $28\times28$ grayscale images for training and 10,000 $28\times28$ grayscale images for testing. We first convert the images into spike trains, the encoded spike patterns are trained with the fully connected SNN and convolutional SNN, respectively. The complete convolutional SNN model are shown in Fig. \ref{CNN model}. Table \ref{table1} shows the classification accuracies of the two SNNs, and other spike-driven learning algorithms on the MNIST dataset.
\newcommand{\tabincell}[2]{\begin{tabular}{@{}#1@{}}#2\end{tabular}}  
\begin{table*}[h]
\normalsize
\centering
\caption{The classification accuracies of  existing spike-driven learning algorithms on the MNIST dataset. We use the following notation to indicate the SNN architecture. Layers are separated by - and spatial dimensions are separated by $\times$. The convolution layer and pooling layer are represented by C and P, respectively.}
\label{table1}
\begin{tabular}{l l l l l}
\hline
Model & Coding & Network Architecture & Additional Strategy & Acc. (\%) \\ \hline
 Mostafa \cite{mostafa2017supervised} & Temporal      & 784-800-10  &      Weight and Gradient Constraintion          &    97.5         \\ 
 
  Tavanaei et al \cite{tavanaei2019bp}   &  Rate      &   784-1000-10                &      None           &        96.6     \\
  
  Comsa et al \cite{comsa2019temporal} &    Temporal    &     784-340-10               &       Weight and Gradient Constraintion          &  97.9           \\ 
   Kheradpisheh et al\cite{kheradpisheh2019s4nn}   &   Temporal     &    784-400-10                &     Weight constraintion             &     97.4        \\ 
 \textbf{STDBP (This work)}  &   Temporal     &      784-400-10              &      None           &     \textbf {98.1}       \\ 
 \textbf{STDBP (This work)}    &    Temporal    &        784-800-10          &    None             &    \textbf{98.5}         \\ 
 \textbf{STDBP (This work)}     &    Temporal    &    \tabincell{c}{28$\times$28-16C5-P2-32C5\\-P2-800-128-10} &    None             &    \textbf{99.4}         \\ 
\end{tabular}
\end{table*}

\begin{figure}[htbp]
\centering 
\subfigure{
\begin{minipage}{9cm}
\includegraphics[scale=0.65]{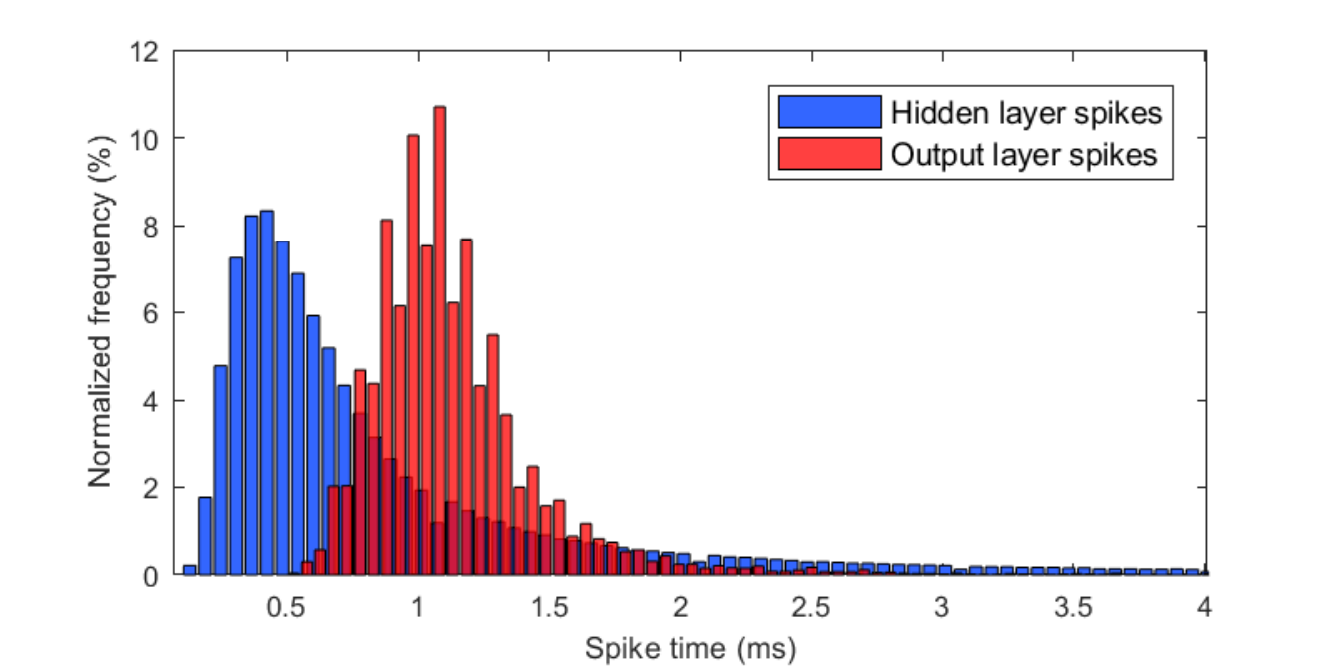} 
\centering 
\caption{(a)}
\end{minipage}
}
\subfigure{ 
\begin{minipage}{9cm}
\centering 
\includegraphics[scale=0.65]{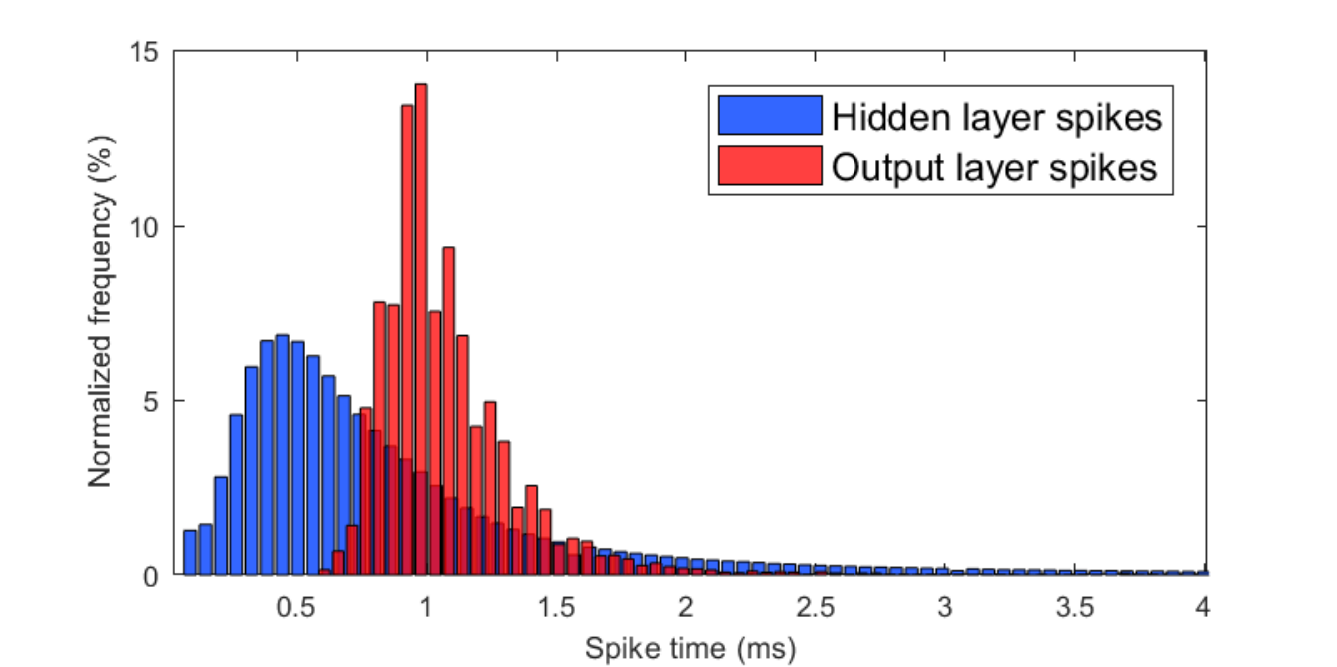} 
\centering 
\caption{(b)}
\end{minipage}
}
\caption{Histograms of spike times in the hidden layers and the output layer across 10000 test images for the two SNNs: (a) 784-400-10 and (b) 784-800-10).} 
\label{fig4} 
\end{figure}

As shown in Table \ref{table1}, the proposed STDBP learning algorithm could reach accuracies of $98.1\%$ and $98.5\%$ with network structures of 784-400-10 and 784-800-10, respectively. They outperform previously reported results with same network structure. For example, with the structure of 784-400-10, the classification accuracy of our method is $98.5\%$, while the accuracy achieved by Mostafa \cite{mostafa2017supervised} is $97.5\%$. Another advantage of our algorithm is that it does not need additional training strategies, such as constraints on weights and gradient normalization, which are widely used in previous works to improve their performance \cite{mostafa2017supervised,comsa2019temporal,kheradpisheh2019s4nn}. This facilitates large-scale implementation of STDBP, and make it possible to train more complex CNN structure. The proposed convolutional SNN achieves an accuracy of is $99.4\%$, much higher than all the results obtained by the fully connected SNNs. To our best knowledge, this is the first implementation of a convolutional SNN structure with single-spike-timing-based supervised learning algorithms.

Fig. \ref{fig4} shows the distribution of spike timing in the hidden layers and of the earliest spike time in the output layer across 10000 test images for two SNNs, namely 784-400-10 and 784-800-10.  For both architectures, the SNN makes a decision after only a fraction of the hidden layer neurons. For the 784-400-10 topology, an output neuron spikes (a class is selected) after only $48.6\%$ of the hidden neurons have spiked. The network is thus able to make very rapid decisions about the input class. In addition, during the simulation time, only 66.3\% of the hidden neurons have spiked. Therefore, the experimental results demonstrate that the proposed learning algorithm works in a accurate, fast and sparse manner.

\begin{figure}[htbp]
\centering 
\subfigure{
\begin{minipage}{9cm}
\centering
\includegraphics[scale=0.62]{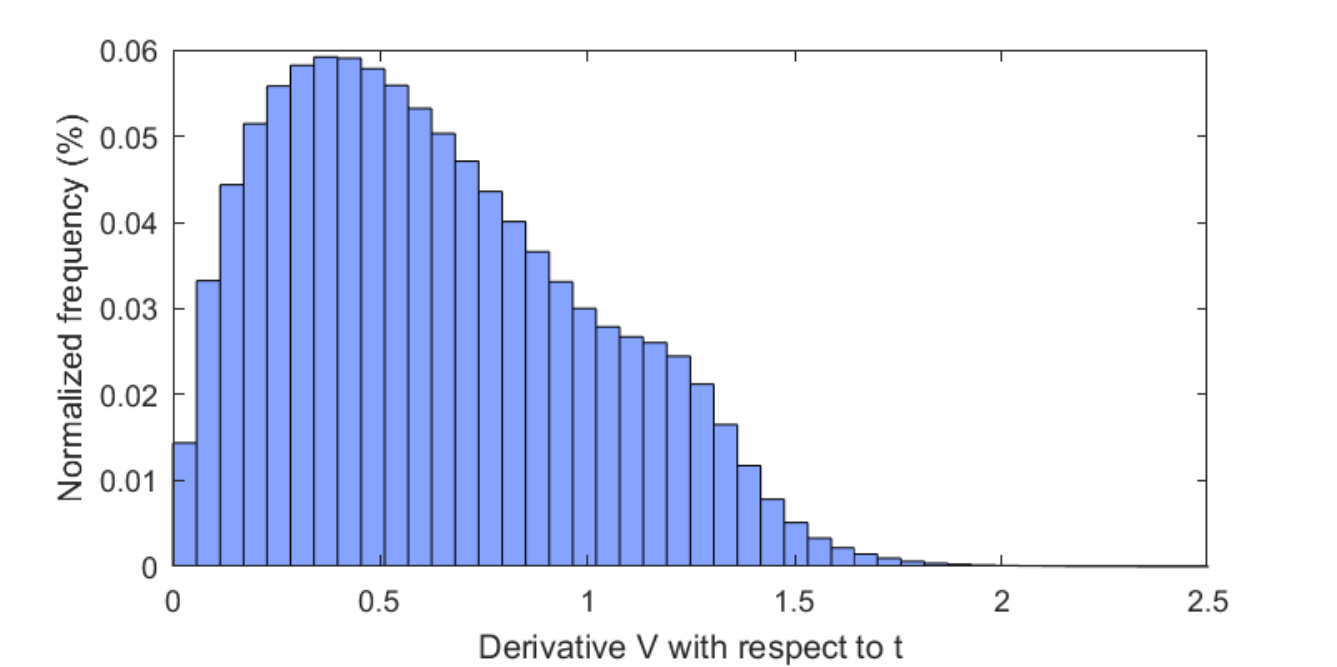} 

\centering
\caption{(a)}
\end{minipage}
}

\subfigure{ 
\begin{minipage}{9cm}
\centering 
\includegraphics[scale=0.62]{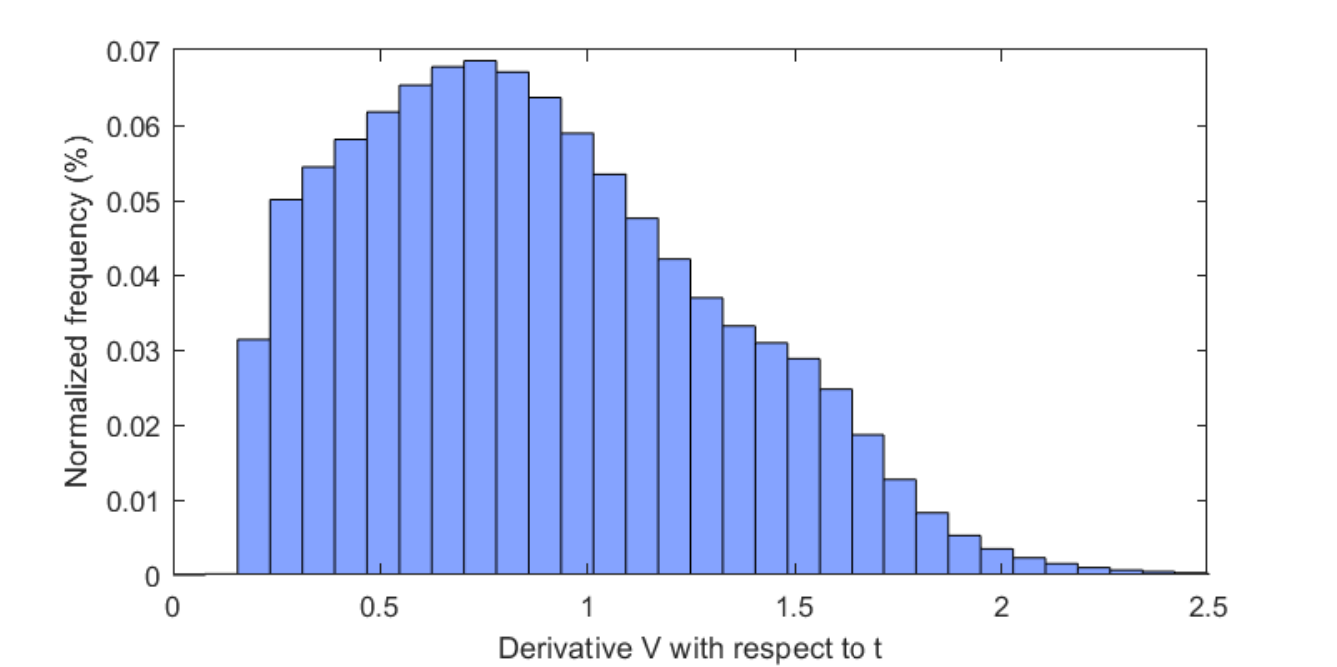} 

\centering
\caption{(b)}
\end{minipage}
}
\centering
\caption{Distribution of spike gradients of different PSP functions. (a)~Alpha-PSP function. (b)~The proposed ReL-PSP function.} 
\label{gradientexplodes} 
\end{figure}

\begin{figure}
    \centering
    \includegraphics[scale=0.60]{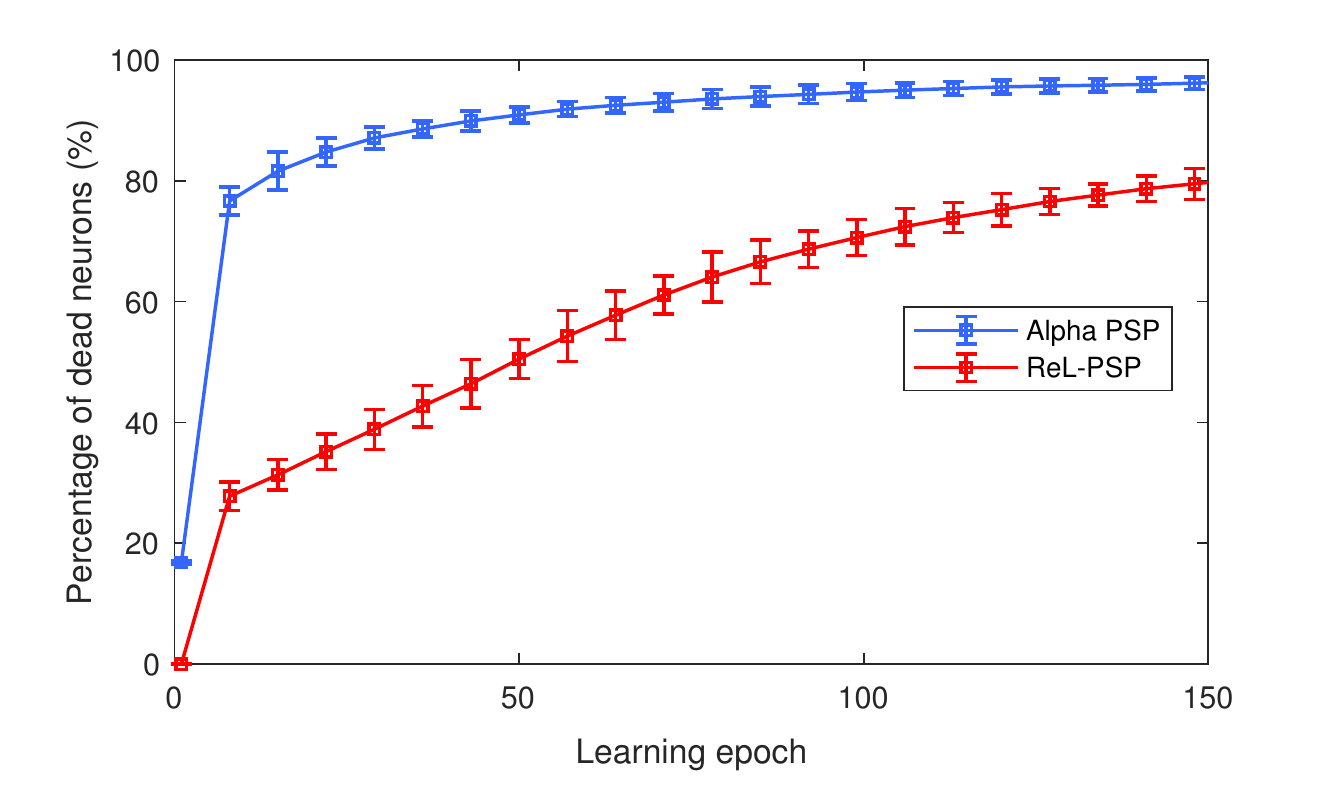}
    \caption{ The number of the dead neurons increases along the learning process between Alpha-PSP and ReL-PSP function. }
    \label{deadneuron}
\end{figure}

To investigate whether the proposed ReL-PSP solves the problems of gradient exploding and dead neuron, we take the fully connected SNNs as an example and 20 independent experiments are conducted with different initial synaptic weights.
 The values of gradient and the number of dead neurons are reported in Fig. \ref{gradientexplodes} and Fig. \ref{deadneuron}.

Fig. \ref{gradientexplodes}(a) and (b) show the distribution of spike gradients of the ReL-PSP and the alpha-PSP function, respectively. As shown in Fig. \ref{gradientexplodes}(a), there are some spike gradients is near 0, which will cause the gradient exploding problem. Previous attempts to alleviate this problem, such as adaptive learning rate \cite{shrestha2015adaptive} and dynamic firing threshold \cite{hong2019training}, can not fully resolve it. Fig. \ref{gradientexplodes}(b) shows the spike gradients distribution of the proposed ReL-PSP function. The values of spike gradients are far away from 0 to prevent the gradient exploding problem. Therefore, stable gradient propagation in DeepSNNs can be achieved by the proposed ReL-PSP function.

 Fig. \ref{deadneuron} shows the trend of the number of dead neurons along the learning process with different postsynaptic potential functions. As learning progresses, the number of the dead neurons increases in both SNNs. In neural networks, typically only a subset of neurons are activated~\cite{glorot2011deep} in action. However, excessive inactivated neurons lead to the dead neuron problem, which reduces effective capacity of the network and affects the predictive performance. As we mentioned in \autoref{sec3}-A, due to the leaky nature of traditional PSP function and the spike generation mechanism, SNNs are more likely to suffer from the dead neuron problem. As shown in Fig.~\ref{deadneuron}, after 100 learning epochs, the SNN with ReL-PSP has $30\%$ active neurons and can achieve an accuracy of $98.4\%$ on the test dataset. However, the SNN with alpha-PSP only has $5\%$ active neuron with an accuracy of $92.5\%$. The experimental results corroborate our hypothesis that traditional PSP functions suffer from dead neuron problem, that limits the network predictive performance.

\subsection{Fashion-MNIST dataset}
Fashion-MNIST dataset \cite{xiao2017fashion} has same number of training/testing samples as MNIST, but it is  more challenging than MNIST. The samples are associated with 10 different classes like T-shirt/top, Pullover, Trouser, Dress, Sandal, Coat, Shirt, Bag,
Sneaker and Ankle boot. In this part, we use it to compare our method with the existing fully-connected feedforward SNNs and convolutional SNNs.

\begin{table*}
\caption{The classification accuracies of existing SNN-based computational models on the Fashion-MNIST dataset.}
\centering
\normalsize
\begin{tabular}{llll}
\hline
Model               & Coding   & Network Architecture          & Acc. (\%)  \\
\hline
S4NN \cite{kheradpisheh2019s4nn}              & Temporal & 784-1000-10                   & 88.0      \\
BS4NN\cite{kheradpisheh2020bs4nn}           & Temporal & 784-1000-10                   & 87.3      \\
Hao et al.\cite{hao2020biologically}   & Rate     & 784-6000-10                   & 85.3      \\
Zhang et al.\cite{zhang2020temporal} & Rate     & 784-400-400-10                & 89.5      \\
Ranjan et al.\cite{ranjan2019novel} & Rate  & \tabincell{c}{28$\times$28-32C3-32C3-P2-128-10} & 89.0 \\
 \textbf{STDBP (This work)} & Temporal     & 784-1000-10 & 88.1  \\
\textbf{STDBP (This work)} & Temporal     & \tabincell{c}{28$\times$28-16C5-P2-32C5-P2-800-128-10} & 90.1 \\ 
\end{tabular}
\label{fashionmnist}
\end{table*}

Table \ref{fashionmnist} shows the classification accuracies and characteristics of different methods on Fashion-MNIST dataset. The proposed learning algorithm still deliver the best test accuracy in temporal coding based SNN methods. For example, our method can achieve accuracies of 88.1\% and 90.1\% with the fully-connected SNN and convolutional SNN, respectively. These results outperforms the best reported result of $88.0\%$ ~\cite{kheradpisheh2019s4nn} in temporal coding based SNN models.

\subsection{Caltech face/motorbike dataset}

\begin{figure}
    \centering
    \includegraphics[scale=0.42]{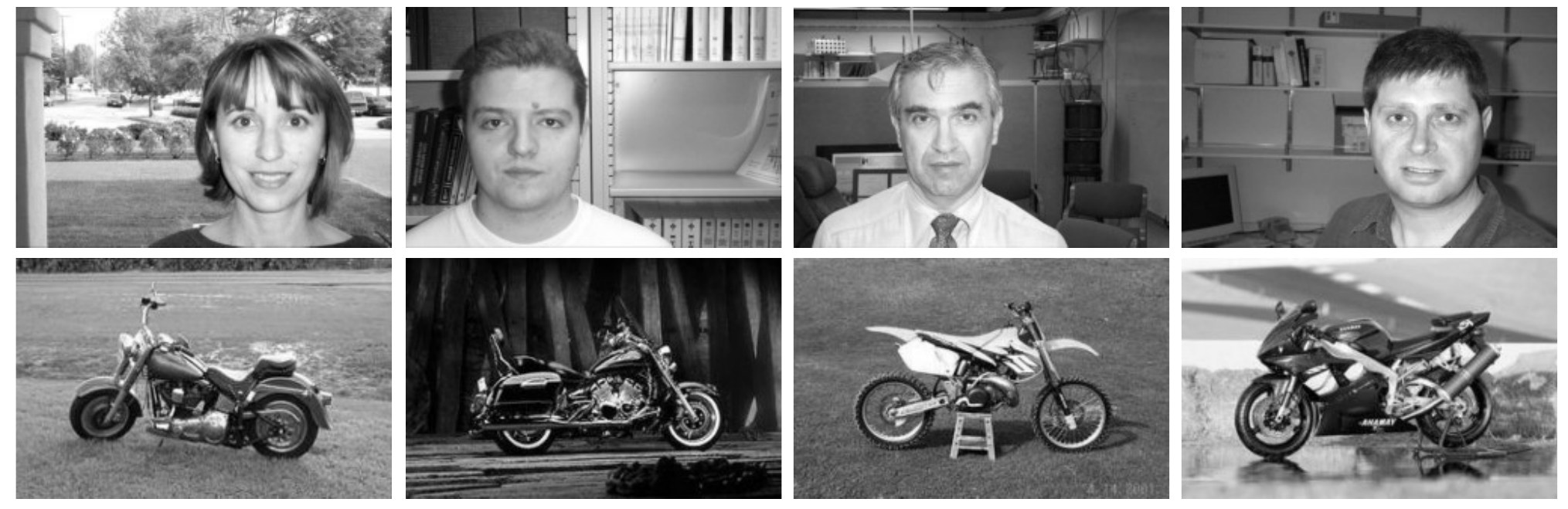}
    \caption{ Some samples of the converted images from Caltech 101 face/motorbike  dataset}
    \label{ImageFigure}
\end{figure}

In this experiment, the performance of the proposed STDBP is evaluated on the face/motobike categories of the Caltech 101 dataset (http://www.vision.caltech.edu).  For each category, the training and validation set consist of 200 and 50 randomly selected samples, respectively, and the others are regard as testing samples. Before training, all images are rescaled and converted to $160\times250$ grayscale images. Fig.~\ref{ImageFigure} shows some samples of the converted images. The converted images are then encoded into spike patterns by temporal coding.

\begin{table*}[h]
\centering
\caption{The classification accuracies of existing SNN-based computational models on the Caltech face/motorbike dataset.}
\normalsize
\label{table2}
\begin{tabular}{l l l l l}
\hline
Model &Learning method &Network Architecture &Classifier &  Acc. (\%) \\ \hline
 Masquelier et al.\cite{masquelier2007unsupervised}      &    Unsupervised STDP  & HMAX \cite{riesenhuber1999hierarchical} & RBF   &    99.2         \\ 
 
Kheradpisheh et al.  \cite{kheradpisheh2018stdp}        &      Unsupervised STDP & \tabincell{c}{28$\times$28-4C5-P7-20C16-P2-10C5}&   SVM               &        99.1    \\
  
  Mozafari et al. \cite{mozafari2018first}    &       Reward modulated STDP     & HMAX \cite{riesenhuber1999hierarchical}   & Spike-based &  98.2           \\ 
   Kheradpisheh et al.\cite{kheradpisheh2019s4nn}      &  Spike-based backpropagation &  $160\times250-4-2$ &  Spike-based               &        99.2        \\ 
 \textbf{STDBP (This work)}   &   Spike-based backpropagation  &  $160\times250-4-2$ &  Spike-based   &  \textbf {99.2}       \\ 
 \textbf{STDBP (This work)}        &   Spike-based backpropagation &  \tabincell{c}{28$\times$28-16C5-P2-32C5-P2-800-128-10} & Spike-based  &      \textbf{99.5}         \\ 
\end{tabular}
\end{table*}

The classification accuracies of different SNN-based computational models are shown in Table~\ref{table2}. The proposed STDBP learning algorithm achieves an accuracy of $99.2\%$ with the fully connected SNN structure and an accuracy of $99.5\%$ with the convolutional SNN structure. The accuracy obtained by STDBP outperforms the previously reported SNN-based methods on this dataset. For example, In Kheradpisheh et al \cite{kheradpisheh2018stdp}, an convolutional SNN structure with a SVM classifier achieves an accuracy of $99.1\%$ on the same dataset. Moreover, it is not a fully spike-based computational model that the membrane potential is used as the classification signal. Recently, a spike-based fully connected SNNs model achieves an accuracy of $99.2\%$. However, the proposed method with convolutional SNN structure reaches an accuracy of $99.5\%$, which is the state-of-the-art performance on this benchmark.

\subsection{Hardware Simulation Results}
\begin{table*}[h]
\centering
\caption{Comparison of a fully connected SNN with various neuromorphic accelerators on the MNIST dataset sorted by accuracy. Acc. is Top-1 accuracy in percent, fps is frames per second, Tech is CMOS technology node in nm, power in mW.(Scaled for 28~nm process ($\times1.17$ for half a generation))}
\normalsize
\label{table4}
\begin{tabular}{l l c r r r r}
\hline
{Accelerator} & Coding. & Acc. (\%) & \multicolumn{1}{c}{fps} & {Tech} & {Power} & {uJ/frame} \\ \hline
SNNwt~\cite{du2015neuromorphic} & Rate & 91.82 & -& 65 & - & 214.700 \\ \hline
TrueNorth-a~\cite{esser2015backpropagation} & Rate & 92.70 & 1000 & 28 & 0.268 & 0.268 \\ \hline
Spinnaker~\cite{khan2008spinnaker} & Rate & 95.01 & 77 & 130 & 300.000 & 3896.000 \\ \hline
Tianji~\cite{ji2018bridge} & Rate & 96.59 & - & 120 & 120.000 & - \\ \hline
Shenjing~\cite{wang2019shenjing} & Rate & 96.11 & 40 & 28 & 1.260 & 38.000 \\ \hline
\bf{STDBP+YOSO (This work)} & Temp. & 98.45 & 21 & 22 & \bf{(0.878*)~0.751} & \bf{(41.93*)~35.839} \\ \hline
TrueNorth-b~\cite{esser2015backpropagation} & Rate & 99.42 & 1000 & 28 & 108.000 & 108.000 \\ \hline
\end{tabular}\\
\end{table*}

We will now focus on evaluating the power and energy efficiency of our SNN models by accelerating their inference operations on YOSO. As a baseline for comparison with other neuromorphic accelerators, we considered a fully connected SNN with network architecture 784-800-10 and trained it on MNIST data with proposed STDBP learning algorithm. The learned weights are then transferred to YOSO for accelerating the inference operations. As shown in Table~\ref{table4}, YOSO consumes 0.751 mW of total power consumption and 47.71 ms of latency to classify an image from MNIST dataset. In addition, YOSO achieves 399$\times$, 159.8$\times$, 143.8$\times$, and 1.67$\times$ power savings as compared to Spinnaker \cite{khan2008spinnaker}, Tianji \cite{ji2018bridge}, TrueNorth-b \cite{esser2015backpropagation}, and Shenjing \cite{wang2019shenjing}, respectively. Though, TrueNorth-a consumes less power than YOSO, there is a significant difference between the classification accuracies of TrueNorth-a and YOSO (See Table~\ref{table4}). Moreover, YOSO provides 108.7$\times$, 6$\times$, and 3$\times$ energy efficiency as compared to Spinnaker \cite{khan2008spinnaker}, SNNwt~\cite{du2015neuromorphic}, and TrueNorth-b \cite{esser2015backpropagation}, respectively.

\section{Discussion and Conclusion}

In this work, we analysed the problems that BP faces in a DeepSNN, namely the non-differentiable spike function, the exploding gradient, and the dead neuron problems. To address these issues, we proposed the Rectified Linear Postsynaptic Potential function (ReL-PSP) for spiking neurons and the STDBP learning algorithm for DeepSNNs. We evaluated the proposed method on both a multi-layer fully connected SNN and a convolutional SNN. Our experiments on MNIST showed an accuracy of $98.5\%$ in the case of the the fully connected SNN and $99.4\%$ with the convolutional SNN, which is the state-of-art in spike-driven learning algorithms for DeepSNNs. 

There have been a number of learning algorithms for DeepSNNs, such as conversion methods  \cite{esser2015backpropagation,hunsberger2015spiking,essera2016convolutional,o2013real,liu2017noisy,diehl2015fast,diehl2016truehappiness,rueckauer2017conversion,rueckauer2018conversion,han2020rmp}, and surrogate gradients methods \cite{wu2018spatio,wu2019direct,neftci2019surrogate,panda2016unsupervised,lee2016training,zenke2018superspike}. These methods are not compatible with temporal coding and spike-based learning mechanism. 
To overcome the non-differentiability of spike function, many methods have been studied \cite{bohte2002error,shrestha2017robust,xu2013supervised,robustnessshrestha2017robustness,hong2019training,mostafa2017supervised,kheradpisheh2019s4nn,comsa2019temporal} to facilitate backpropagation. Two common drawback of these methods are exploding gradients and dead neurons, which have been partially addressed using techniques such as constraints on weights and gradient normalization. These techniques affect learning efficiency, thus limiting the scope of their applications in large-scale networks. The proposed STDBP learning algorithm with ReL-PSP spiking neuron model can train DeepSNNs directly without any additional technique, hence allowing the DeepSNNs to scale, as shown in the high accuracy of the convolutional SNN. 

In addition to being fast, sparse and more accurate, the proposed ReL-PSP neuron model and STDBP have other attributes that might make it more energy-efficient and (neuromorphic) hardware friendly. Firstly, the linear ReL-PSP function is simpler than alpha-PSP for hardware implementation. Secondly, unlike rate-based encoding methods that require more time to generate enough output spikes for classification, our method takes advantage of temporal coding and uses a single spike, which is more sparse and energy-efficient, given energy is mainly consumed during spike generation and transmission. Thirdly, without additional training techniques, on-chip training in neuromorphic chips would be much easier to realize. Finally, even if the training is performed offline, inference of our SNN models can be accelerated with much less power and energy consumption as compared to other state-of-the-art neuromorphic accelerators. 

\bibliographystyle{IEEEtran} 
 
\bibliography{myref}

\end{document}